**An Approach to Grounding AI Model Evaluations in Human-derived Criteria**

Sasha Mitts

Meta FAIR, sashamitts@meta.com

In the rapidly evolving field of artificial intelligence (AI), traditional benchmarks can fall short in attempting to capture the nuanced capabilities of AI models. We focus on the case of physical world modeling and propose a novel approach to augment existing benchmarks with human-derived evaluation criteria, aiming to enhance the interpretability and applicability of model behaviors. Grounding our study in the Perception Test and OpenEQA benchmarks, we conducted in-depth interviews and large-scale surveys to identify key cognitive skills — such as Prioritization, Memorizing, Discerning, and Contextualizing — that are critical for both AI and human reasoning. Our findings reveal that participants perceive AI as lacking in interpretive and empathetic skills, yet hold high expectations for AI performance. By integrating insights from our findings into benchmark design, we offer a framework for developing more human-aligned means of defining and measuring progress. This work underscores the importance of user-centered evaluation in AI development, providing actionable guidelines for researchers and practitioners aiming to align AI capabilities with human cognitive processes. Our approach both enhances current benchmarking practices and sets the stage for future advancements in AI model evaluation.

## 1 INTRODUCTION

Progress in AI research today is evaluated primarily by computable, established benchmarks. These benchmarks provide a standardized measure of progress, essential for efficient and predictable model development. Progress on common benchmarks, such as Large Vocabulary Instance Segmentation (LVIS) and Common Objects in Context (COCO) for object segmentation, or Perception Test and OpenEQA for video scene reasoning, is commonly regarded as a useful indicator of capability improvement [20]. However, these benchmarks can encounter limitations. They can become saturated, where scores plateau or become less meaningful despite model improvements [13, 27]. Additionally, benchmarks may yield results that are difficult to interpret, with scores increasing without clear insights into what aspects of the system have improved [9, 14]. Finally, benchmark scores may not reliably predict real-world or complex scene performance [24]. To address these challenges, we propose augmenting traditional benchmarks with human-derived evaluation criteria. Our approach aims to enhance interpretability and applicability, increasing real-world relevancy. Although developed within the context of physical world modeling benchmarks, our methods can be applied to various research domains [4].

## 2 METHODS

This work focuses on enhancing existing benchmarks rather than creating new ones. Perception Test was selected as a pilot benchmark for this process. Perception Test is designed to assess the perception and reasoning abilities of multimodal video models [1, 22]. Perception Test includes tasks such as object and point tracking, action and sound localization, and video question-answering, testing visual, audio, and text modalities, involving skills like memory, abstract pattern recognition, physics, and semantics. We concentrated on multiple-choice video question-answering, sampling 792 videos and selecting six that represented three question types: Object Shuffle (similar to the 'shell game'), Action Count (tracking the frequency of actions), and Character Order (recalling a sequence of letters). These were chosen to satisfy our experimental design and simplify pilot experimentation. Each question had three response options, with one correct answer (viewable in Appendix 1). In comparison, OpenEQA is a benchmark designed to test if model understanding of an environment (input via video) is sufficient to answer questions about said environment in natural language [18, 21]. Questions test seven categories of reasoning abilities, of which we select three (one video for each of functional reasoning, spatial understanding, and world knowledge, viewable in Appendix 2), where humans were shown to have comparatively weak absolute performance and the delta between humans and the top scoring model (GPT-4V) was small [6].

## 2.1 Experimental Design and Setup

This study consisted of two phases. We first conducted in-depth interviews (n=8, 60 minutes) to identify the reasoning skills essential for answering Perception Test questions [11, 23]. Participants recruited for introspective ability answered multiple-choice questions and explained their reasoning process, detailing how they arrived at their answers and what made their approach effective. Subsequently, we revealed an AI's responses to the same questions and asked participants to evaluate these against their own. They shared assumptions about the AI's evaluation process and suggested improvements.

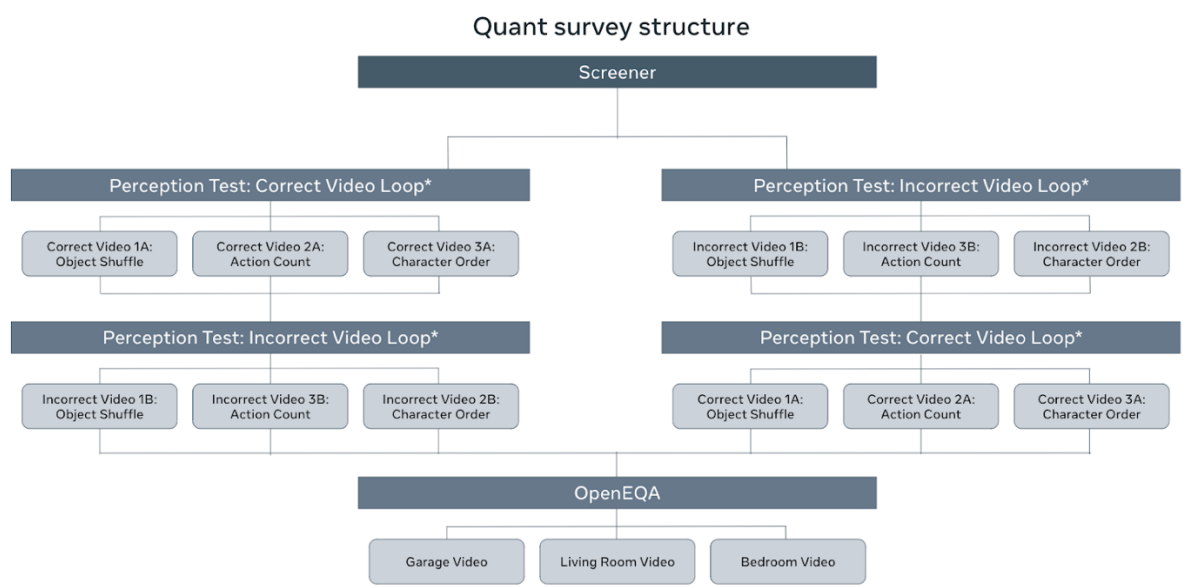

Figure 1. Survey Design

Interview findings were coded and analyzed to identify key skills used in reasoning over physical world modeling tasks. We identified six skills and corresponding questions were constructed to test their perceived importance via a survey (n=1,515), Figure 1, (questions in Appendix 3). Participants rated the importance of each skill for solving Perception Test and OpenEQA questions and assessed their confidence in their answers before and after viewing a model's responses. Participants also rated the perceived accuracy and impressiveness of the model, as well as the contribution of each skill to the AI's performance. The order of correct/incorrect was randomized by respondent; respondents evaluated only one correct and one incorrect video in the Perception Test portion of the survey. Respondents did not evaluate the same type of video, i.e., one correct Object Shuffle and one incorrect Object Shuffle. The results of the interviews and survey were synthesized to further identify and prioritize the skills that contribute to the physical world modeling capabilities tested, as well as to propose questions to evaluate progress in terms of these skills as part of or to inform future benchmarks (Figure 2).

## 3   RESULTS

In the context of Perception Test, participants identified Prioritization (76%, T2B) and Memorizing (77%, T2B) as the most critical for arriving at correct answers, significantly above average compared to each of the other six skills (Figure 3). When reflecting on OpenEQA, participants emphasized the comparative importance of Discerning (83%, top-two box) and Contextualizing (80%, top-two box), (Figure 4). These skills were deemed crucial for both open-ended and close-ended tasks, suggesting that developing benchmarks to track progress in these areas will help train AI models to perform physical world modeling tasks in alignment with user expectations. Participants noted that skills like Interpreting, Discerning, and Empathizing were particularly important for AI compared to human reasoning in physical world modeling. This perception may stem from participants viewing AI as more literal and formulaic. Interestingly, no skills were rated as more important for humans, possibly because participants considered human intelligence the baseline. Participants had high expectations for AI performance, often assuming that even incorrect answers were accurate [17]. When they disagreed with the AI's responses, they sometimes doubted their own judgment or rationalized the AI's reasoning. This trust appeared to be based on perceptions of AI's strengths, such as being free from distractions and capable of processing vast amounts of information. However, there is room to improve participants' trust in AI performance. Summaries of AI's reasoning could help users understand how AI approaches tasks, thereby boosting confidence in outputs [10, 17, 19].

| Most important | | Important | | Less important | |
|---|---|---|---|---|---|
| **Prioritizing** | **Memorizing** | **Interpreting** | **Discerning** | **Contextualizing** | **Empathizing** |
| Knowing the relative significance of actions when they happen | Recording relevant information in an accurate manner and being able to recall it | Understanding the intent of a question, parsing any explicit or implicit meanings | Recognizing distinctions between objects by noting their differences | Referencing memory to establish constraints or make inferences, rather than for facts | Understanding the challenge from another perspective |
| Rated as *equally important* for humans and the model:<br>• **76%** rated as very or extremely important **for them** (T2B)<br>• **76%** rated as very or extremely important **for AI** (T2B) | Rated as *equally important* for humans and the model:<br>• **75%** rated as very or extremely important **for them** (T2B)<br>• **77%** rated as very or extremely important **for AI** (T2B) | Rated as *more important* for the model:<br>• **67%** rated as very or extremely important **for them** (T2B)<br>• **72%** rated as very or extremely important **for AI** (T2B) | Rated as *more important* for the model:<br>• **63%** rated as very or extremely important **for them** (T2B)<br>• **72%** rated as very or extremely important **for AI** (T2B) | Rated as *equally important* for humans and the model:<br>• **62%** rated as very or extremely important **for them** (T2B)<br>• **64%** rated as very or extremely important **for AI** (T2B) | Rated as *more important* for the model:<br>• **58%** rated as very or extremely important **for them** (T2B)<br>• **65%** rated as very or extremely important **for AI** (T2B) |
| **INGOING PERCEPTIONS:** | | | | | |
| *AI is "smart" enough and has seen enough data to know what's important* | *AI has perfect memory because it's a machine* | *AI is literal:* AI takes things at face value, whereas humans read between the lines<br><br>"It interprets the question the way I phrase the question." –47 y/o male | *AI is formulaic:* AI is only as smart as its training data, whereas humans can make informed deductions<br><br>"AI mistook two pens for being the same." –25 y/o female | *AI can easily identify patterns to inform its decision-making* | *AI is robotic:* AI does not have feelings, whereas humans can care for each other<br><br>"The human element [is missing], like emotions." –31 y/o male |

November 2024, Panel Data
Participant Benchmark Importance. How important were each of the following for you to solve the task? | AI Benchmark Importance. How important do you think each of the following were for the AI to solve the task?
n = 1515 (based on only first loop viewing)

Figure 2. Skills Findings Overview

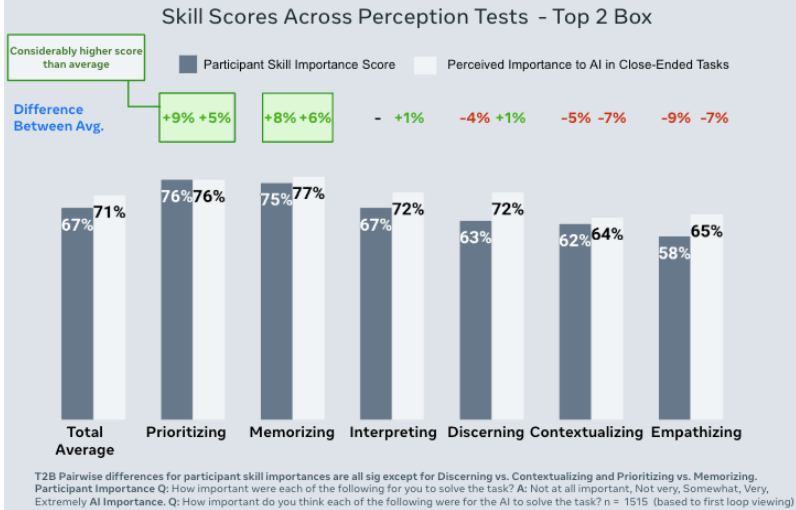

Figure 3. Perception Test – Perceived Skill Importance

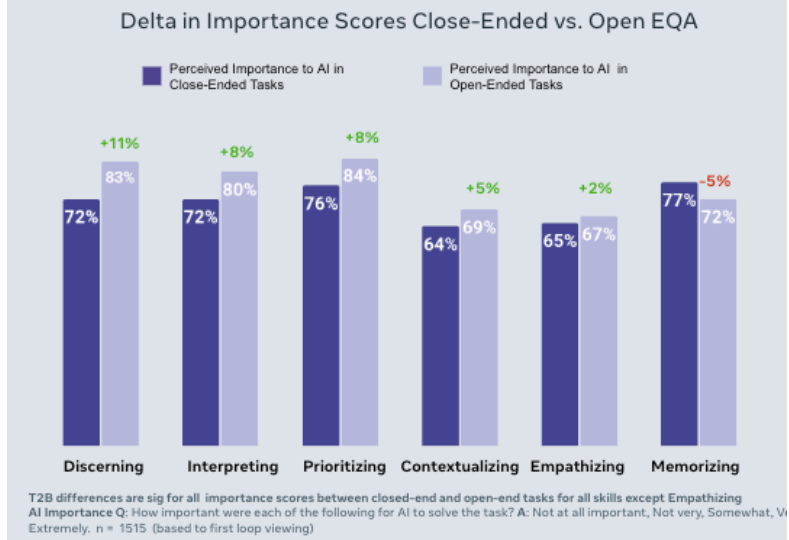

Figure 4. Perception Test Versus OpenEQA – Perceived Skill Importance

## 3.1 Skills Deep-dive

We identified micro-skills that contributed to the skills of Prioritization and Memorizing, with the aim of improving AI performance in physical world modeling tasks (viewable in Appendix 4). For Prioritization, core components included Focus, Reprioritization, and Distraction Filtering. These skills were tested through questions that assessed the temporal, spatial, and semantic relevance of a scene, the ability to update priorities dynamically, and the establishment of thresholds for ignoring irrelevant stimuli. For Memorizing, key components included Temporal Recall, Source Attribution, and Recall Application. Questions targeting these skills evaluated the allocation of attention across time intervals, the salience and relevance of information, and the ability to establish relationships between information and outcomes.

The language and questions developed for these skills can be integrated into existing benchmarks or used as tools for constructing new benchmarks. This ensures that the skills identified and scaled deliberately reflect the cognitive processes humans use and expect AI to deploy in physical world modeling tasks. By focusing on these nuanced and meta-cognitive micro-skills, we propose questions to serve as or steer evaluation criteria which are potentially less prone to saturation, decompose performance into more interpretable segments, and align with how users expect AI to apply physical world modeling skills. Future work will formalize these evaluation axes into validated questions to augment benchmarks like Perception Test and OpenEQA, and demonstrate domain adaptability.

## 4 DISCUSSION

Beyond the specific results of our study, we outline a process for leveraging validated benchmarks to create human-grounded evaluation measures for AI models. This process can guide others in developing similar criteria across various domains.

1. Align with Research Goals: Collaborate with research scientists to understand the overarching goals of your domain. Determine what models under development aim to solve or deliver, beyond specific product outcomes.
2. Assess Current Metrics: Identify how progress is currently measured. Often, metrics are applied in a fragmented manner. Filter to those that are widely respected and relied upon.
3. Evaluate Metric Trustworthiness: Understand why these metrics are considered trustworthy. This could be due to the benchmark's diversity, scientific rigor, or lack of alternatives.
4. Select High-Impact Benchmarks: Finalize selection of benchmarks that effectively track progress toward significant goals.
5. Analyze Model Responses: Examine the types of responses models provide on these benchmarks and the variance across different runs.
6. Conduct Human Evaluations: Establish an interview guide for in-depth, moderated human completion of selected benchmark tasks. Recruit participants who are patient, expressive, and capable of introspection.
7. Analyze Human Responses: Code and analyze participant responses to identify how they decompose their reasoning steps and align with benchmark goals.
8. Survey Skill Importance: Develop a survey to test the relative importance of identified skills. Allow participants to express which skills they perceive as most consequential for both humans and AI.
9. Quantify Skill Importance: Use survey data to quantitatively assess the importance of each skill and identify informative correlations.
10. Propose Benchmark Improvements: Suggest refinements to the benchmarks based on your findings, and explore ways to generalize these improvements to other benchmarks.

This structured approach ensures that human-grounded measures are integrated into AI model evaluations, aligning with user expectations and enhancing model design.

### 4.1 Limitations and Next Steps

As we move forward, several practical steps are necessary to refine our approach. First, a comprehensive literature review will help map the skills identified in physical world modeling to existing knowledge in cognitive science and decision theory. This will clarify what constitutes reasoning and whether current AI benchmarks adequately test it.

Our approach is intentionally grounded in existing benchmarks, which provides a practical starting point for improvement. However, alternative methodologies could explore cognitive processes from a broader perspective, potentially offering foundational insights into how tasks like physical world modeling are approached cognitively.

It's important to acknowledge that our method is optimized for enhancing current processes rather than defining cognitive processes universally. This focus ensures that our findings are relevant to the specific context and needs of our team. Finally, we acknowledge the benefits of recruiting from a larger and more diverse sample for the qualitative portion of this work, but find the present sample sufficient to produce a proof of concept which can be tested.

## 5 CONCLUSION

We hope this overview serves as a practical guide for teams working on physical world modeling or other AI benchmarking efforts. By integrating user research with model evaluation, we aim to ensure that downstream needs and expectations are considered early in model design. This collaboration between user experience research and modeling teams fosters a deeper understanding of the decisions and thought processes involved in AI development to serve human needs.

### ACKNOWLEDGMENTS

We wish to acknowledge the contributions of the team at FAIR working on physical world modeling in characterizing benchmarking approaches worth investigating, and the support of Now What Research in data collection for this study.

**APPENDICES**

**A1. Perception Test Questions Used**

| Test | Video | Question | Answer | Model Response | Options |
|---|---|---|---|---|---|
| **Object Shuffle** | video_4324.mov | The person uses multiple similar objects to play an occlusion game. Where is the hidden object at the end of the game from the person's point of view? | **(c) Under the third object from the left** | (a) Under the second object from the left | (a) Under the second object from the left, (b) under the first object from the left (c) under the third object from the left |
| **Action Count** | video_1803.mp4 | How many times did the person show objects to the camera? | **(a) 7** | (b) 4 | (a) 7, (b) 4, (c) 5 |
| **Character Order** | video_9824.mp4 | What letter is likely to be shown next? | **(a) O** | (c) E | (a) O, (b) L, (c) E |
| **Object Shuffle** | video_215.mp4 | The person uses multiple similar objects to play an occlusion game. Where is the hidden object at the end of the game from the person's point of view? | (a) Under the third object from the left | (a) Under the third object from the left | (a) Under the third object from the left, (b) under the first object from the left (c) under the second object from the left |
| **Action Count** | video_6927.mp4 | How many times did the person launch objects on the table? | (c) 4 | (a) 2 | (a) 2, (b) 9, (c) 4 |
| *Character Order* | video_4232.mp4 | *What was the order of the letters on the table before shuffling?* | *(b) abcdef* | *(c) dacbef* | *(a) fdacbe, (b) abcdef, (c) dacbef* |

**A2. OpenEQA Questions Used**

<u>Garage - Planning - functional reasoning:</u> ID: 87019892-7a7f-4ce0-b1e6-4c0d6e87b90a

[25] Q: What can I use to water my plants?

[25] A: The green hose

<u>Bedroom - Visual and multimodal reasoning - world knowledge:</u> ID: 1015cab7-827a-437c-ab86-6e4c2a9dd083

[40] Q: What style of paintings are put up in the bedrrom [sic]?

[40] A: Abstract

<u>Living Room - Spatio-temporal Memory - Spatial Understanding:</u> ID: 884e2405-85f6-446f-a0ab-a03e537e5184

[46] Q: What is in the center of the room with the stairs?

[46] A: A red couch

**A3. Survey Skill Question Language Example**

**Benchmark Importance.** How important were each of the following for you to identify how many objects were launched?

**COLUMNS**

1. Not at all important
2. Not very important
3. Somewhat important
4. Very important
5. Extremely important

**ROWS**

**[INTERPRETING]**

1. Interpreting what the question is really asking (e.g., not taking it at face value)

**[DISCERNING]**

2. Discerning the difference between objects (e.g., knowing the difference the ball and the binder)

**[EMPATHIZING]**

3. Understanding from another's perspective (e.g., taking into account multiple ways to solve the question)

**[CONTEXTUALIZING]**

4.     Building on past experiences to answer the question (e.g., understanding physics)

**[PRIORITIZING]**

5.     Knowing which moments to pay attention to (e.g., focusing on the end result of the ball)

**[MEMORIZING]**

6.     Remembering beginning vs. end state (e.g., memorizing the number of attempts)

**A4. Micro-skills details**

| Prioritization micro skills | Focus — Concentrating attention and mental effort on specific information and maintaining sustained engagement | Reprioritization — Adjusting the order of priorities based on new information, changing circumstances, or evolving goals | Distraction Filtering — Ignoring or blocking out irrelevant stimuli or interruptions in order to focus on the most important information |
|---|---|---|---|
| Questions to assess performance on prioritization micro skills | How do you decide which moments are most critical to focus on when presented with a series of events? | How did you handle situations where multiple events or actions in the video were competing for attention? | Did you consider which information was less relevant to the task at hand? |
| | How did you determine which pieces of information were essential to the task at hand? | How did you handle changes in priorities during the task when new information or circumstances arose? | How did you manage distractions or interruptions during the task assigned? |
| | Which parts of the video were most important in helping you answer the question and why? | How did you decide when it was time to adjust your priorities or approach to the task given? | How did you block out external interruptions (e.g., extraneous details) while completing the task? |

Prioritization *micro-skill*s

| Memorization micro skills | Temporal Recall — Remembering the initial state of information and how it evolved or concluded | Source Attribution — Distinguishing between using provided information and relying on past experience to formulate a response | Recall Application — Applying remembered information to effectively answer questions based on past content or experiences |
|---|---|---|---|
| Questions to assess performance on memorization micro skills | Were you able to remember beginning and end states by the time you responded? | Did you formulate your response based on the information shared or did you rely on past experience to come up with an answer? | How did you use the information you knew prior to this video to answer the question about this video? |
| | How did you identify, understand, and interpret the significance of changes in information over time? | How did you decide when it was appropriate to reference prior conversations or stick strictly to new input? | Which pieces of information provided in the task did you apply directly to your solution? |
| | How did you navigate if there was conflicting information from earlier and later parts of the same question, task, or challenge? | How did you integrate prior knowledge with new information without causing confusion? | How did the information given in the task help you arrive at your final answer? |

Memorization *micro-skill*s

| Discernment micro skills | Nuance Identification — Identifying and differentiating subtle variations between objects to make distinctions | Classification — Distinguishing and organizing different categories of objects effectively | Resolving Visual Ambiguity — Navigating unclear visual information, like blurry objects or rapid movements |
|---|---|---|---|
| Questions to assess performance on discernment micro skills | Were there any subtle differences between the objects that you had to recognize in order to distinguish them? | How did you determine which objects belonged in which category while working on the task? | How did you prioritize certain visual cues over others when interpreting an image? |
| | How did you ensure you weren't overlooking minor details that could affect your answer? | What criteria were applied to distinguish between these categories? | When faced with visual elements that were difficult to interpret, what methods did you use to clarify their meaning? |
| | How did you compare and contrast objects with very similar characteristics? | What process did you follow to redefine categories if new information suggested changes were needed? | When faced with conflicting visual signals, how did you decide which one to trust to resolve the ambiguity? |

Discernment micro-skills

| Interpretation micro skills | Inferential Reasoning — The ability to draw conclusions and make assumptions based on implicit cues or incomplete information | Meaning Decoding — Deciphering the true meaning of a question, especially when its wording might be misleading or unclear | Sequential Interpretation — Using prior information or steps in a conversation to decipher what the question is asking |
|---|---|---|---|
| Questions to assess performance on interpretation micro skills | Did you consider any hidden assumptions or implications in the way the question was phrased? | How did you approach understanding a question when the wording was potentially ambiguous? | How did you use the context of the conversation (or the previous questions) to interpret what the question was really asking? |
| | How did you draw conclusions when the information provided was incomplete or implied rather than explicitly stated? | How did you determine the most relevant meaning of a question based on context? | How did you use the sequence of information leading up to the question to ensure you understood it correctly? |
| | When confronted with incomplete details, how did you determine what was most likely or reasonable to assume? | How did you identify the parts of the question that could be misinterpreted and correct them? | Did you integrate earlier steps or pieces of information in the conversation to help you interpret the meaning of the current question? |

Interpretation micro-skills